\documentclass[letterpaper, 10 pt, conference]{ieeeconf}  

\IEEEoverridecommandlockouts                              

\usepackage{graphicx}
\usepackage{todonotes}
\usepackage{soul}
\usepackage{tabularx}

\title{\LARGE \bf Shaping Expressiveness in Robotics: The Role of Design Tools in Crafting Embodied Robot Movements}

\author{Elisabetta Zibetti$^{1}$, Alexandra Mercader$^{2}$, Hélène Duval$^{3}$, Florent Levillain$^{4}$, Audrey Rochette$^{3}$\\ and David St-Onge$^{2}$
\thanks{$^{1}$Elisabetta Zibetti is with CHArt Laboratory, Department of Psychology,
        University Paris8, Paris, France
        {\tt\small ezibetti@univ-paris8.fr}}%
\thanks{$^{2}$Alexandra Mercader and David St-Onge are with the Lab of INIT Robots, Department of Mechanical Engineering, École de technologie supérieure,
        Montréal, Canada
        {\tt\small name.surname@etsmtl.ca}}%
\thanks{$^{3}$Hélène Duval and Audrey Rochette are with the Department of Dance, Université du Québec à Montréal,
        Montréal, Canada
        {\tt\small surname.name@uqam.ca}}%
\thanks{$^{4}$Florent Levillain is with Costech Laboratory, Department of Humanities and Social Sciences,
        Université technologique de Compiègne, Compiègne, France
        {\tt\small florent.levillain@utc.fr}}%
}

\begin{document}

\maketitle
\thispagestyle{empty}
\pagestyle{empty}

\begin{abstract}
As robots increasingly become part of shared human spaces, their movements must transcend basic functionality by incorporating expressive qualities to enhance engagement and communication. This paper introduces a movement-centered design pedagogy designed to support engineers in creating expressive robotic arm movements. Through a hands-on interactive workshop informed by interdisciplinary methodologies, participants explored various creative possibilities, generating valuable insights into expressive motion design. The iterative approach proposed integrates analytical frameworks from dance, enabling designers to examine motion through dynamic and embodied dimensions. A custom manual remote controller facilitates interactive, real-time manipulation of the robotic arm, while dedicated animation software supports visualization, detailed motion sequencing, and precise parameter control. Qualitative analysis of this interactive design process reveals that the proposed "toolbox" effectively bridges the gap between human intent and robotic expressiveness resulting in more intuitive and engaging expressive robotic arm movements.
\end{abstract}

\section{Introduction}
As robots become prevalent within environments shared with humans, the need to design robot motions that go beyond functionality to include expressiveness and intuitive readability has become increasingly critical. Expressive motion design has emerged as an effective approach for conveying emotions and enhancing socially meaningful human-robot interactions \cite{Hagane2022}.

A variety of methods have been investigated to address the challenge of expressive robot motion design, including movement analysis frameworks such as Laban notation \cite{Burton2016}, zoo-anthropomorphic approaches \cite{sauer2021}, and biomimetic strategies inspired by biological systems \cite{naoko2022}. Advanced 3D animation tools with temporal editing features enable precise control over expressive timing \cite{Moioli2022}, while recent research leveraging Large Language Models (LLMs) has reproduced robot movements comparable to those created by professional animators \cite{mahadevan2024}. Nevertheless, significant gaps remain in methods and tools that facilitate expressive motion design, particularly for engineers unfamiliar with expressive movement principles. The primary challenges include the absence of straightforward methods for translating expressive intentions into robotic motion parameters and the difficulty of transferring virtual animations onto physical robots due to inherent mechanical constraints, such as weight, inertia, and joint limitations. These challenges frequently necessitate labor-intensive manual adjustments, limiting creative exploration.

\begin{figure}[t]
    \centering
    \includegraphics[width=\columnwidth]{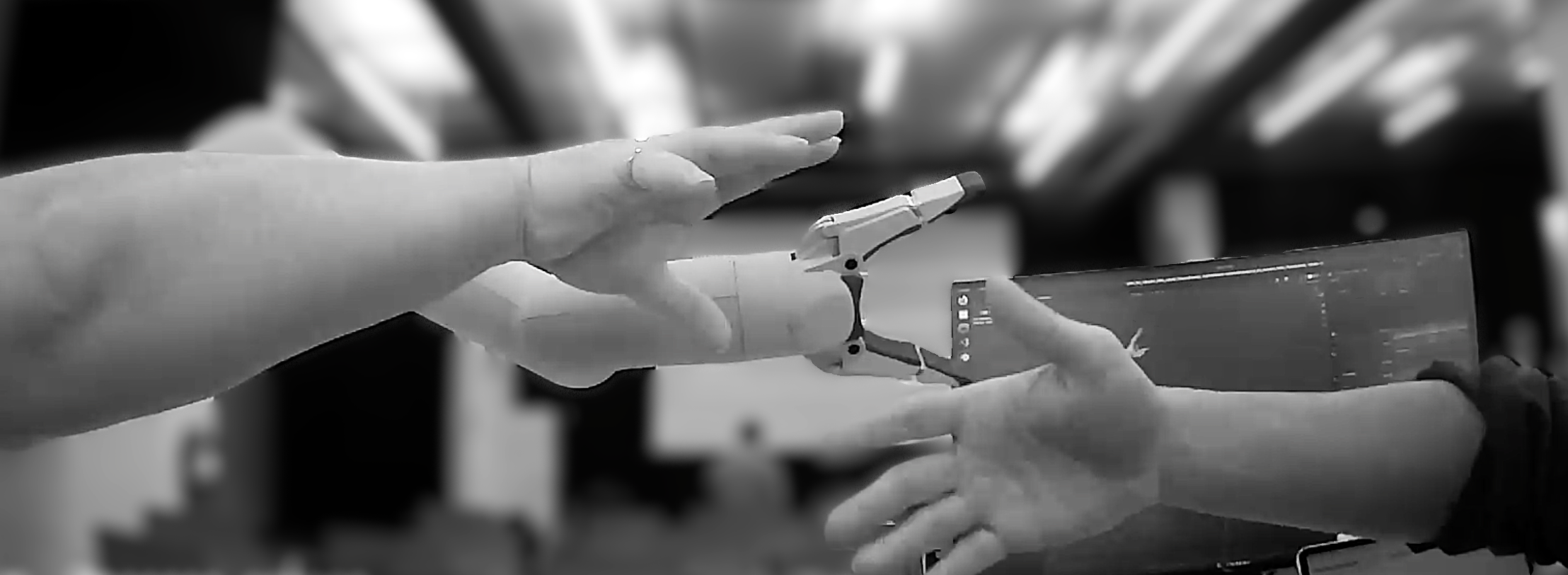}
    \caption{Two participants composing a movement for a robotic arm as part of our workshop.}
    \label{fig:armtouch}
    \vspace{-1.95em}
\end{figure}

Inspired by remote controllers and intuitive input devices commonly used in 3D animation software (e.g., Blender), this paper proposes a movement-centric design pedagogy emphasizing physical engagement and embodied interaction. Our approach leverages humans' innate sensitivity to motion, promoting intuitive interactions that enable engineer-designers to engage seamlessly with digital design tools. Instead of solely using traditional inputs such as keyboards and mice, we explore physical and motion-based interfaces alongside graphical representations to better align with natural human perception and movement intuition.

Our methodology centers on "embodied interaction," defined as the creation, manipulation, and communication of meaning through active physical engagement with artifacts \cite{pepperell2003}. It also recognizes that tangible interaction inherently takes place within physical spaces \cite{Hornecker2006}. We integrate embodied movement concepts from dance and choreography \cite{Warburton2011} into engineering contexts. Although engineers typically excel at technical implementations, they often lack the conceptual frameworks and perceptual sensitivity required to articulate and manipulate movement expressiveness effectively \cite{wendy2007}. In contrast, dance and choreography offer established frameworks to interpret movement not merely as functional displacement, but as inherently expressive and communicative. By introducing engineers to dynamic movement qualities, kinesthetic empathy, and intentionality, we provide analytical tools for discerning nuanced aspects of robotic movements that convey an agent’s attitude toward tasks or environments \cite{knight2017}. Additionally, our embodied-interactive motion design toolbox benefits from constructionist and problem-based learning approaches, fostering creative integration of both technical and conceptual components into the design process \cite{chen2020}.

To evaluate our proposed toolbox, we conducted a workshop in which participants utilized customized tools to collaboratively address a design challenge: creating expressive robotic arm movement sequences tailored to specific scenarios. A complementary research question guided this exploration: How do tools featuring different interaction modalities (e.g., direct manipulation versus graphical representations) help designers develop intuitive insights into movement qualities and their expressive potential? During the workshop, participants' interactions with each tool were recorded, and semi-structured interviews captured explicit reflections on their experiences. Verbatim transcripts were subsequently analyzed qualitatively to identify the strategies, challenges, and advantages participants encountered when shaping expressive robotic movements.

In the following sections, we present the theoretical background and the developed toolbox, which aims to bridge engineers' expressive intentions with the practical generation of expressive robot movements, supported by a creative and engaging design process.

\section{Related works}
\subsection{Balancing Functionality and Expressiveness}

Creating expressive robotic movements requires balancing functional objectives with expressive intentions across diverse social contexts. Engineering disciplines typically exhibit a technical-social dualism, prioritizing technical elements over social and expressive considerations \cite{wendy2007}. Individual creativity manifests uniquely according to cognitive styles, such as analytical versus intuitive, visual versus verbal, or preference for novelty versus tradition \cite{Roebuck2020}. Consequently, design tools differ significantly in their accessibility and intuitiveness for different users. Conventional tools such as joysticks and animation software vary in technical engagement levels, while conceptual frameworks like Movement Observation-Analysis (MOA) \cite{harbonnier2021} provide foundational tools for crafting expressive animations reflective of human movement diversity. Moreover, tangible manipulation, spatial interaction, embodied facilitation, and expressive representation enhance collaborative knowledge-building among designers \cite{Hornecker2006}, while experimentation with diverse tools fosters mutual inspiration, prompting designers to view design aspects from multiple perspectives \cite{SCHAPER2023}.

Constructionist learning theory, proposed by Papert \cite{Papert1993}, emphasizes that individuals learn most effectively through active engagement in creating personally meaningful artifacts \cite{Gaudiello2013}. Central to this theory are ‘objects-to-think-with’—artifacts bridging embodied experiences with abstract concepts, enabling tangible and shareable understanding. Interactive tools align naturally with this approach by providing immediate feedback and promoting active, creative participation rather than passive use \cite{Resnick1996}.

\subsection{Conceptual Tools for Movement Exploration}\label{sec:moa}
Movement Observation-Analysis (MOA) is a qualitative approach \cite{harbonnier2021}, integrating Laban/Bartenieff Movement Analysis (LBMA) with the Functional Analysis of the Dancing Body (AFCMD). LBMA provides structured frameworks and vocabularies for articulating movement qualities and their relationships, significantly influencing robotic expressiveness studies \cite{venturereview2019, Burton2016}. AFCMD \cite{harbonnier2023}, emphasizes body awareness, proprioception, and a functional understanding of movement through somatic practices.

MOA integrates and enriches these approaches, offering a holistic perspective that unifies the functional and expressive dimensions of movement \cite{harbonnier2021}. Applied to robotics, MOA encourages designers to consider robotic agents comprehensively, acknowledging mechanical capabilities as inherently expressive. The dynamic qualities central to MOA, based on Laban’s Effort Theory (Tab. \ref{tab:laban}), serve as crucial conceptual tools for designing and analyzing robotic expressivity. MOA categorizes these dynamic qualities into four tonalities: spatial (unidirectional or multidirectional), temporal (accelerated or decelerated), weight (light, strong, or heavy), and flow (controlled or free). Spatial tonality can be unidirectional—focused on a specific point—or multidirectional, spreading attention across a broader field. Temporal tonality can manifest as acceleration or deceleration. Weight tonality encompasses sensations of lightness, strength (muscular effort relative to gravity), or heaviness. Flow tonality characterizes the movement's restraint or release, distinguishing between controlled and free motions.

\begin{table}[]
\centering
\caption{Mobilized MOA parameters (\emph{LBMA’s analogue})}
\vspace{-1em}
\begin{tabularx}{0.9\columnwidth}{|>{\setlength\hsize{\hsize}\setlength\linewidth{\hsize}}X|>{\setlength\hsize{\hsize}\setlength\linewidth{\hsize}}X|}
\hline{\textbf{Spatial tonality (\emph{space})}} & {\textbf{Temporal tonality (\emph{time})}} \\ \hline
\begin{itemize}
    \item Unidirectional (\emph{direct})
    \item Multidirectional (\emph{indirect})
\end{itemize}                  & \begin{itemize}
    \item Accelerated (\emph{sudden})
    \item Decelerated (\emph{sustained})
\end{itemize} 
    \\ \hline
    {\textbf{Flow Tonality (\emph{flow})}}    & \multicolumn{1}{l|}{\textbf{Weight tonality (\emph{weight})}}   \\ \hline
\begin{itemize}
    \item Controlled (\emph{bound})
    \item Unhindered (\emph{free})
\end{itemize}                            & \begin{itemize}
    \item Light (\emph{light})
    \item Strong (\emph{strong})
    \item Heavy  
\end{itemize}   \\
\hline
\end{tabularx}
    \label{tab:laban}
\vspace{-1.85em}
\end{table}



While the MOA system enhances the understanding of expressiveness in movement, it is not intended as a direct compositional tool. Rather, it serves as a guide for shaping intuition and intention through a structured, iterative process consisting of three core steps:

\textbf{Subjective Impressions}: Observers begin by allowing kinesthetic impressions to emerge—what the movement might feel like in their own body—eliciting emotional and intentional responses through embodied empathy.

\textbf{Movement Parameter Analysis}: These initial impressions are then examined using MOA’s specialized vocabulary of movement parameters. This analytical step allows observers to articulate, support, or refine their perceptions with greater precision.

\textbf{Construction of Meaning}: Finally, observers assess how the identified parameters interact within the given context. If the expressive intent appears inconsistent or unclear, the process is repeated to reassess and refine both the analysis and interpretation.

This cycle creates a feedback loop between subjective experience and formal analysis, underscoring that expressive movement emerges from the interplay of intuitive perception and structured observation~\cite{harbonnier2021}. Although MOA provides a detailed conceptual framework that can inform expressive robotic design, its practical application in robotics is non-trivial. Effective translation of MOA principles into robotic movement requires (i) a technically appropriate and creatively flexible design toolkit, and (ii) a pedagogical framework that enables designers to integrate these principles meaningfully into practice.

\subsection{Robotic arm trajectory planning}
Expressive robotic motion, involving meticulous control over velocity, acceleration, and jerk \cite{Mahzoon2022}, requires effective trajectory planning. Trajectory planning methodologies and commercial solutions typically utilize either joint-space or Cartesian-space paradigms. Joint-space planning offers joint-level precision but may produce unintuitive end-effector trajectories, potentially resulting in inefficient or unpredictable movements \cite{MALVIDOFRESNILLO2023}. Conversely, Cartesian-space planning focuses on end-effector paths, providing smoother, more intuitive motions but relying on complex inverse kinematics models, which introduce limitations such as singularities and obscure underlying joint configurations \cite{MALVIDOFRESNILLO2023}.

Both joint-space and Cartesian-space methods can incorporate high-level techniques like optimization-based approaches and sampling strategies to enhance planning efficiency and accuracy. These methods explore configuration spaces to generate collision-free, smooth paths while optimizing constraints such as energy efficiency and joint limit compliance \cite{MALVIDOFRESNILLO2023, montazer2022}. Alternatively, predefined trajectories created by robotics experts can aim at replicating naturalistic movements, including animal behaviors \cite{sauer2021} and anthropomorphic gestures \cite{knight2016}.

Most robotic arms include remote controllers (teach pendants) supporting Cartesian or joint-space control, and occasionally advanced interfaces such as a 6D mouse enabling simultaneous multi-degree-of-freedom manipulation. Nevertheless, conventional solutions remain subject to the mentioned limitations. The most intuitive current teaching solution involves admittance control, which dynamically adjusts robot stiffness and compliance based on operator-applied forces, enabling direct physical manipulation \cite{LABRECQUE2018}. However, admittance control demands advanced embedded sensing, limiting portability across different robotic platforms.

\section{Design and Implementation of the Toolbox}

The proposed motion design toolbox, framed by a pedagogical approach, sits at the intersection of various choreographic methodologies to facilitate the creation of expressive robotic movements \cite{laviers_choreographic_2018}. It emphasizes body-based practices that foster kinesthetic empathy \cite{strukus_mining_2011}, coupled with embodied interaction, enabling spontaneous and intuitive manipulation of robotic arms. This approach encourages participants to understand the robot's constraints and possibilities, incorporate expressive movement qualities (see Tab. \ref{tab:laban}), and experience interactions within the robot's environment. Additionally, the toolbox leverages graphical representations inspired by choreographic tools to aid designers in visualizing and refining animated sequences \cite{calvert_evolution_1993}. Thus, it effectively combines embodied insights \cite{fdili_alaoui_strategies_2015} with technical practices.

\subsection{Enhance participants’ awareness with MOA}

Our first goal is to enhance participants' kinesthetic awareness and comprehension of expressive movement parameters using a structured three-step process aligned with MOA methodology. (1) Initially, expressive movements are presented as co-creations, influenced by both the robot's actions and observers' interpretations. Given this subjective dynamic, kinesthetic empathy becomes vital for understanding and interpreting motions. (2) Participants then observe predefined robotic sequences, sharing their spontaneous impressions, consistent with MOA's initial observation stage. (3) Subsequently, they are introduced explicitly to MOA’s dynamic qualities (see Sec. \ref{sec:moa}) to support analytical observation and facilitate the creation of expressive sequences. These steps help participants interpret, design, and appreciate movements by focusing their attention on specific dynamic qualities and structuring their analyses.

\subsection{Motion Design Workflow}

\begin{figure}[h]
    \centering
    \includegraphics[width=\columnwidth]{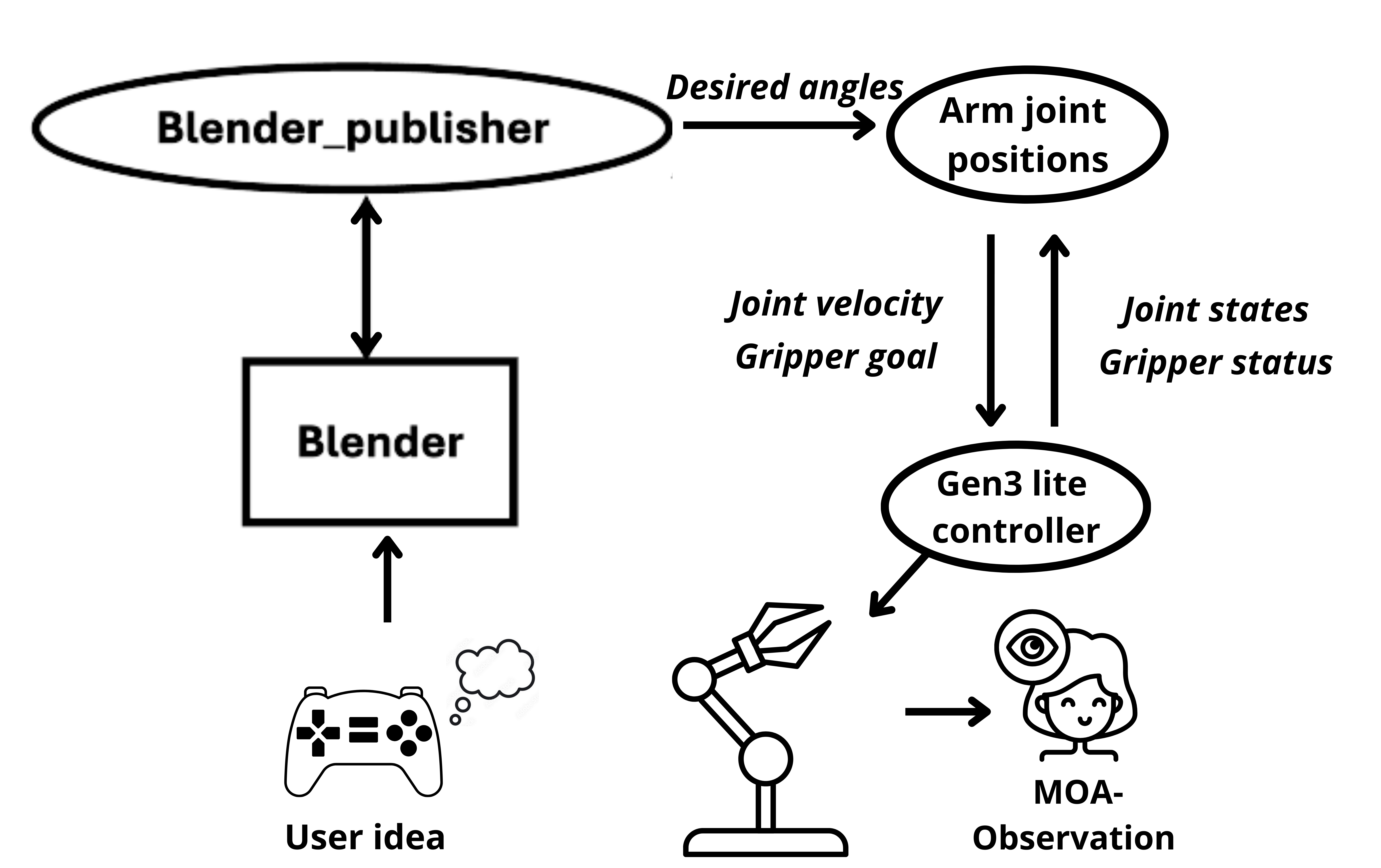}
    \caption{Workflow diagram illustrating the transfer of user ideas into robotic motions via Blender and real-time ROS communication.}
    \label{fig:DesignwithMOA}
    \vspace{-1em}
\end{figure}

The iterative process of transforming a designer's conceptual ideas into robotic motions is illustrated in Fig. \ref{fig:DesignwithMOA}. This workflow involves initial real-time motion conceptualization via direct manipulation, followed by refinement and execution within Blender.

Throughout this iterative design cycle, participants are invited to observe and analyze the designed motion using the MOA three steps analytical process across direct manipulation, graphical animation, and final robotic execution. This process could ensure that dynamic tonalities remain consistent with the designer's creative intentions.

In Sec. \ref{sec:ps4} and \ref{sec:blender}, we detail technological tools developed within our embodied interaction approach, empowering participants to effectively engage with and embody MOA’s conceptual framework for expressive movements.\footnote{The complete software infrastructure and hardware setup for the Kinova Gen3 lite robotic arms are publicly available, including Docker containers for simplified deployment on Windows, macOS, and Ubuntu: https://git.initrobots.ca/jmazerolle/blender-animator.}

\subsection{Remote Controller for Robotic Motion Design}\label{sec:ps4}

To foster embodied interaction with robotic systems and allow intuitive control of all potential movements, we developed a customized PS4 remote controller. Understanding the robot’s capabilities and limitations requires practical engagement beyond what high-level path-planning software alone can offer. Although impedance control methods are intuitive, they typically require advanced and costly robotics not commonly accessible in educational settings. Therefore, we adopted a teleoperation approach compatible with more affordable educational robots, enhancing portability and accessibility across various robotic platforms.

The controller's multiple buttons enable users to switch effortlessly between Cartesian and joint-space control, even during movement\footnote{The workshop material is also available online: https://git.initrobots.ca/expressivity/blender-animator.}. Users can dynamically adjust the velocity independently in each mode. Pre-recorded postures and fault-clearing routines are included to assist users in managing singularities and recovering from operational faults quickly. The ergonomic mapping of translations and rotations has been iteratively refined through extensive use in undergraduate mechanical engineering labs.

Using the remote controller to select joints and adjust their speed enables precise timing and dynamic control, directly shaping movement intention. The seamless toggling between Cartesian and joint modes supports flexible and detailed motion design. Additionally, an inertia toggle and precise analog stick orientation control further enhance expressive movement, allowing sharp, energetic actions or smooth, deliberate gestures.

Most new users become proficient with this remote within 5-15 minutes. However, fine motor skills may limit extended or complex sequences, prompting users to transition to animation software for detailed refinements.

\subsection{Animation for Timing, Smoothness and Preciseness }\label{sec:blender}

\begin{figure}[h]
    \centering
    \includegraphics[width=0.49\columnwidth]{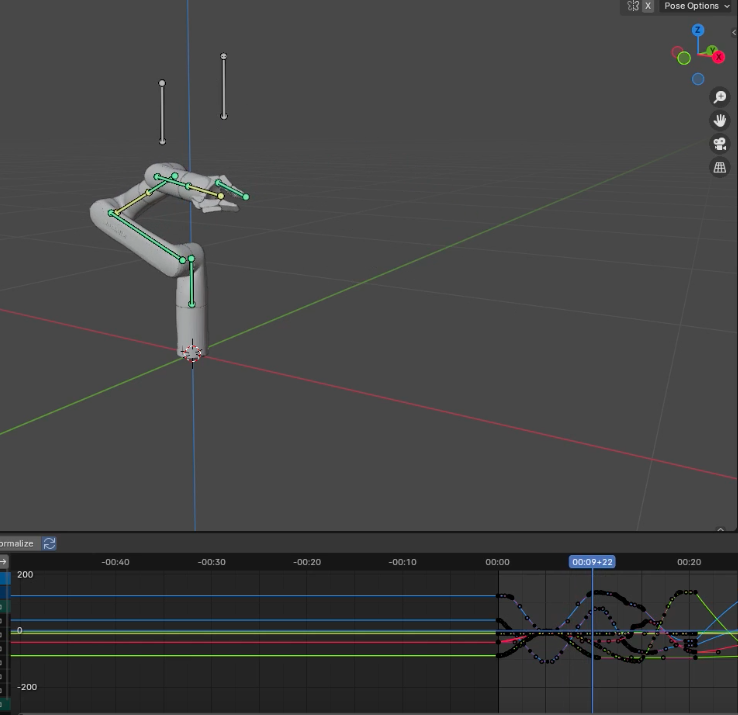}
    \includegraphics[width=0.49\columnwidth]{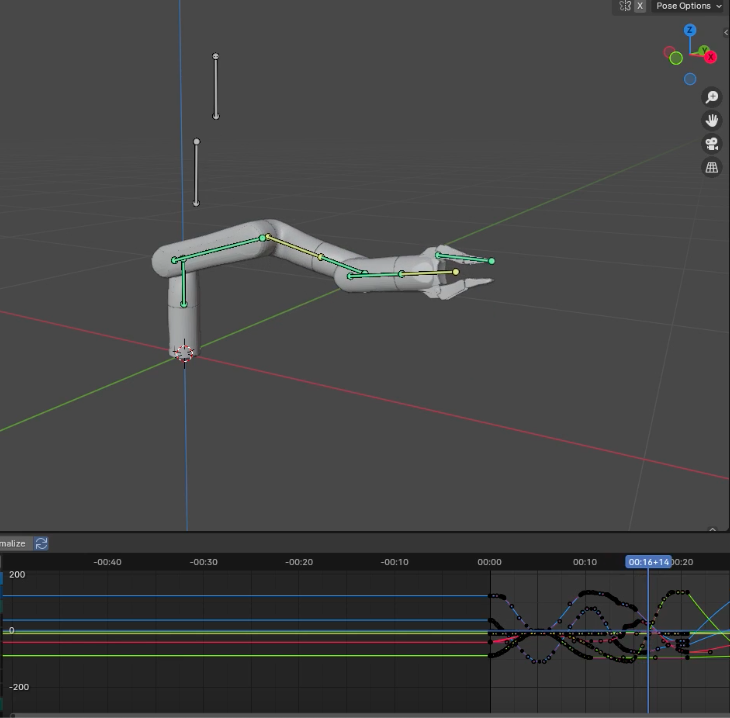}
    \caption{Designing robotic movements in Blender: avatar model and timeline view.}
    \label{fig:blender_dive}
    \vspace{-1.5em}
\end{figure}

We adopted Blender, an open-source animation software, to support precise animation design. By importing the robot arm’s model and its physical characteristics, Blender allows comprehensive visualization and precise control of robotic movements. Its interface (Fig. \ref{fig:blender_dive}) includes a detailed timeline for adjusting joint trajectories and motion smoothness.

Fig. \ref{fig:blender_dive} exemplifies a robotic arm executing a diving motion, utilizing the graphical editor to refine timing and smoothness of joint movements. The graphic editor at the bottom of the Blender interface displays a timeline where each joint’s movement is plotted over time. This timeline is essential for adjusting the time scale and sequence length, as well as refining the smoothness of joint trajectories by manipulating the control points of each joint’s curve. Blender’s interface also allows for the inclusion of more complex motions, such as those involving the robot's elbow and gripper. By carefully adjusting the joint angles and their corresponding timing within the sequence, designers can create nuanced movements that can be repeated across the timeline or saved for future use.

Once the sequence is finalized, Blender transmits real-time commands via the ZeroMQ protocol to a ROS node that translates these into executable robotic commands. A Proportional-Derivative (PD) controller ensures accurate and smooth movement tracking, continuously monitoring and correcting joint states and gripper actions. Our Blender plugin and ROS controller node enable seamless integration, providing a robust environment for designing, testing, and executing sophisticated robotic movements.

\section{User study Protocol}
The primary aim of this three-and-a-half-hour workshop was to examine how participants engaged with various technical tools (robotic arm, remote controller, animation software), their understanding of each tool's unique capabilities, and how these capabilities informed strategies to enhance the expressiveness of robotic motions. Additionally, we explored how different operational modes and types of visual feedback provided by these tools influenced participants’ ability to create expressive movement sequences. Finally, we aimed to evaluate how introducing participants to the conceptual framework of MOA influenced their intuitive understanding and appreciation of expressive movement qualities.

\subsection{Participants}
Twenty-one participants, all with engineering backgrounds and without prior experience in motion design or movement analysis, volunteered for this study. They were organized into eight teams of two or three members, each team randomly assigned to one of two scenarios: (T)herapeutic or (E)mergency (defined below). Before the design session began, participants gave informed consent to be filmed and interviewed during the workshop. The École de technologie supérieure’s ethics committee approved the study (H20220301).
 
\subsection{Tasks and Materials}
The workshop, conducted during a summer school, provided a hands-on and interactive environment for exploration and experimentation. Participants were first introduced to the concept of expressiveness, the importance of movement lisibility, kinesthetic empathy and the MOA parameters. Throughout the session, participants were encouraged to: 1) define specific roles within their groups to streamline their design processes, 2) explore the robotic arm’s capabilities using the remote controller to develop initial movement ideas, 3) familiarize themselves with the animation software and motion recording tools, and 4) collaboratively create expressive movement sequences corresponding to their assigned scenario.

\textbf{Therapeutic Context}:
Participants' goal was to create movements conveying warmth, comfort, and empathy to patients. Teams T1, T2, T5, and T6 worked specifically within this context.

\textbf{Public Safety and Emergency Response}:
Participants' goal was to create movements suitable for assisting in crisis situations such as evacuations or disaster responses. They had to develop motions that clearly communicated urgency and guided individuals toward safety. Teams E3, E4, E7, and E8 worked specifically within this scenario.
   
Each team worked with a Kinova Gen3Lite robotic arm, selected for its accessibility in educational environments and suitability for expressive motion design tasks. Participants utilized a custom-mapped PS4 remote controller (see Section \ref{sec:ps4}) to manually explore and control the arm’s motion. Blender, an open-source 3D animation software (see Section \ref{sec:blender}), allowed further refinement, visualization, and precise sequencing of movements.

\subsection{Data collection}
Qualitative data were gathered through audiovisual recordings (researchers' chest-mounted cameras) and post-session questionnaires. Throughout the session, two interviewers documented participants’ design processes, decisions, and interactions with the tools. Interviews were semi-structured, focusing on four primary areas:

\textbf{Intentions}: What specific outcomes or movements are you aiming to achieve for your assigned scenario?

\textbf{Process}: How are you approaching the design of your movement sequence? Are you following a clear plan or improvising? Which tool seems most appropriate for your goals - why?

\textbf{Current Actions}: What actions are you currently performing? How is the tool you are using helping you achieve your objectives?

\textbf{Challenges}: What difficulties have you encountered, and what might help overcome these challenges?

At the end of the session, each team presented their finalized movement sequence to the group. After these presentations, participants completed a reflective questionnaire, specifically addressing how they applied MOA’s dynamic qualities, identifying the most relevant factors for their scenario, and describing strategies they developed to overcome limitations posed by the robotic system.

\section{Qualitative Analysis}
The semi-structured interviews were transcribed and analyzed in two stages. Initially, we employed an inductive approach inspired by grounded theory \cite{Strauss1994}, allowing themes to organically emerge from participants’ experiences. This phase involved two independent coders who first established the coding framework collaboratively, then separately coded the transcripts, finally discussing their coding results to achieve consensus. Subsequently, a complementary deductive analysis was conducted to identify major thematic categories.

\subsection{Expressive intentions and tool-specific strategies}
When designing expressive movements, participants frequently referenced specific gestures or actions they aimed to emulate. Team T6, working on the therapeutic scenario, described their intention as creating "an understanding, nodding behavior," while a participant from E8, addressing the emergency scenario, stated, "we want to make a sign, like a real human, that says clearly: ‘Go there, evacuate.’

Beyond particular gestures, several participants conceptualized broader behavioral qualities or character traits. Team T1 sought movements that were "funny and with personality," whereas a T5 participant envisioned subtle emotional qualities: "I was thinking something like the robot is shy, so it turns away a little bit."

The specific affordances and constraints of the provided tools notably influenced participants’ design approaches, especially concerning the balance between intuitive control and precision. A participant from T2 directly compared the remote controller and animation software: "With the joystick, you have an intuitive interface, but it's difficult to be precise." Conversely, they highlighted Blender’s contrasting advantage: "With Blender, you see the lines and can directly grab them for greater precision—but what's bad about precision is that it lacks a human touch."

Despite the intuitiveness of the controller, participants faced challenges when composing detailed movement sequences. In joint-space mode, the necessity for precise control combined with the complexity of mapping movements proved cumbersome. An E7 participant reflected, "It feels robotic having to move joint by joint, rather than moving multiple joints simultaneously."

While Cartesian-space mode allowed for more fluid movements, its focus primarily on the gripper resulted in limited awareness of the robot’s overall posture. As the same E7 participant observed, "When you move in space, it's hard to see what the robot does. Yes, you know how the gripper moves, but what happens with the rest of the robot?"

\subsection{Modes of visualizing and interacting with movements}
Unlike the remote controller, Blender emphasizes a distinctly spatial visualization approach. By representing movements graphically, participants perceived sequences as structured wholes with clearly editable segments. A participant from T2 commented, "You really have access to all the joints in a much more visual way than with the controller because you see the lines, you just grab it, and you know."

Conversely, the remote controller encourages a more temporal interaction, where movement design unfolds dynamically and improvisationally. A T1 participant characterized the process as "trial and error," while a T5 participant emphasized the relational aspect: "I was just trying to translate with the joystick that feeling of movement into the arm, one movement at a time."

When participants aimed for complex expressiveness, they frequently turned to Blender to break down sequences, control multiple joints simultaneously, and refine timing and smoothness. A T1 participant described Blender’s advantages: "You can actually take a piece of movement that you created, you can duplicate it, you can move it.... You can copy-paste things, or speed them up, or slow them down.... You have lots of options."

Distinct strategies emerged from using Blender alone or in combination with the controller:

\textbf{Targeted editing}: Adjusting specific segments to better align with expressive intentions and refining joint configurations for smoother trajectories.

\textbf{Iterative refinement}: Exploring and refining multiple variations to progressively enhance expressiveness.

\textbf{Sequence duplication}: Replicating and modifying motion segments to create related yet varied movements.

\textbf{Modular composition}: Breaking complex sequences into smaller, manageable segments for independent editing before reassembling them.

\subsection{Post-workshop questionnaire}

Five of the eight teams completed the post-workshop questionnaire, designed to elicit reflections on: (1) their application of MOA’s dynamic factors, (2) the factors most relevant to their expressive goals, and (3) strategies employed to manage scenario requirements and robotic constraints. Thematic analysis \cite{Braun01012006} of these responses identified several key insights:

\textbf{Impact of MOA’s dynamic factors on motion design:}
The MOA dynamic factors showed varied influence on participants’ design processes. Teams T1 and T6 found these factors helpful, noting they provided structure ("It structured the thought process") and encouraged consideration of diverse movement dimensions. Team T5 specifically cited the concepts of flow and sustain as beneficial in achieving fluidity. Conversely, teams E3 and E7 reported minimal impact, attributing this primarily to limited comprehension of MOA concepts: "It didn't really help, as we didn't understand them well enough" (E7).

\textbf{Relevant factors for expressive design:}
Regarding factors most relevant to their scenarios, T1 emphasized the ‘Direct/Indirect’ factor as crucial for conveying a humorous and lively personality. T6 and E7 highlighted ‘speed and continuity,’ with T6 additionally mentioning ‘space and time.’ Team T5 identified software limitations alongside fluidity as essential to expressing empathetic gestures. In contrast, E3 did not single out any specific factor as particularly relevant to their objective of conveying urgency clearly.

\textbf{Team strategies for addressing scenarios and managing constraints:}
Teams employed various strategies to adapt their sequences and address robotic limitations effectively. T1 focused on recognizable patterns, such as designing sinusoidal trajectories for ease of motion creation and timing adjustments. Teams E3 and E7 initially utilized the remote controller for basic movements, subsequently refining these sequences within Blender for improved fluidity and precision. Specifically, E3 employed Blender for simultaneous joint control. Team T5 emphasized creating slow, flowing movements via the remote, later refined in Blender due to hardware and software constraints. Lastly, T6 explored motions directly from "the robot’s perspective," ensuring feasible trajectories with the controller, then using Blender to fine-tune control and enhance motion smoothness.

The tools’ specific characteristics tended to foster complementary activities in the design process. Participants typically began by determining expressive intentions through their own physical execution of the movement. They then used the remote controller to explore the robot’s capabilities and limitations, improvising sequences that align with their initial ideas. A participant from  E7 described this progressive workflow: “To design the movement, I think we started with what we wanted to do physically (i.e., using the joystick), and then we went to the robot, played with the robot, saw what it was capable of, and now we’re going to record and modify everything in Blender.”

\section{Discussion and Conclusion}

Our study introduced a design toolbox combining a conceptual framework (MOA) and technological tools (custom remote controller and Blender software) to facilitate expressive robotic motion design, integrating dance-inspired embodied concepts into engineering education.

Qualititative analysis highlighted distinct contributions from each tool each of them encouring  different design strategies. The remote controller provided direct and intuitive manipulation of the robot, fostering improvisation and embodied exploration, helping participants develop an embodied understanding of the robot’s capabilities and limitations. 
Conversely, Blender supported detailed analytical refinement through graphical visualization, enabling precise control of sequences. These complementary affordances promoted different yet synergistic strategies, balancing spontaneous experimentation with systematic fine-tuning.

The MOA conceptual framework impacted teams variably; some found it valuable for structuring their creative process and articulating movement qualities, while others integrated it minimally. Direct hands-on interaction frequently led participants to unexpected creative insights, illustrating the value of iterative physical engagement.

Designing expressive robotic motion is more of an art than an engineering challenge. The development of appropriate tools to address the nuanced qualities of expressiveness remains an important area for further research. Future directions include enhancing tool accessibility through simplified interfaces or adaptive controls catering to diverse user expertise and abilities. Strengthening theoretical-practical integration—specifically connecting MOA dynamic factors to practical motion control—is another critical area. Further research might explore how different technological resources influence user perception and manipulation of movement qualities. By refining these tools and methods, future work can expand expressive robotic design capabilities and reinforce human-centered educational practices.

\vspace{-0.6em}
\section{Acknowledgments}
The workshop is possible thanks to the commitment of Jean Mazerolle, Ali Imran, Zakary Kamal Ismail and Alexandre Lapointe. 
\vspace{-1em}
\bibliographystyle{IEEEtran}
\bibliography{expmot,florent}


\end{document}